\documentclass[accepted]{uai2021} 

\usepackage[american]{babel}

\usepackage{natbib} 
\usepackage{mathtools} 
\usepackage{booktabs} 
\usepackage{tikz} 
\usepackage{pbox}
\usepackage{graphicx}
\usepackage{amsmath}
\usepackage{booktabs}
\usepackage{subcaption}
\usepackage{float}

\makeatletter
\newcommand*{\addFileDependency}[1]{
  \typeout{(#1)}
  \@addtofilelist{#1}
  \IfFileExists{#1}{}{\typeout{No file #1.}}
}
\makeatother

\newcommand*{\myexternaldocument}[1]{%
    \externaldocument{#1}%
    \addFileDependency{#1.tex}%
    \addFileDependency{#1.aux}%
}
\usepackage{xr}
\myexternaldocument{supplementary}

\title{Class Balancing GAN with a Classifier in the Loop}

\author[1]{\href{mailto:Harsh Rangwani <harshr@iisc.ac.in>?Subject=Your UAI 2021 paper}{Harsh Rangwani}{}} 
\author[2]{Konda Reddy Mopuri}
\author[1]{R. Venkatesh Babu}

\affil[1]{%
    Indian Institute of Science, Bengaluru
}
\affil[2]{%
    Indian Institute of Technology Tirupati
}

\begin{document}
\maketitle

\begin{abstract}
Generative Adversarial Networks (GANs) have swiftly evolved to imitate increasingly complex image distributions. However, majority of the developments focus on performance of GANs on balanced datasets. We find that the existing GANs and their training regimes which work well on balanced datasets fail to be effective in case of imbalanced (i.e. long-tailed) datasets. In this work we introduce a novel theoretically motivated Class Balancing regularizer for training GANs. Our regularizer makes use of the knowledge from a pre-trained classifier to ensure balanced learning of all the classes in the dataset. This is achieved via modelling the effective class frequency based on the exponential forgetting observed in neural networks and encouraging the GAN to focus on underrepresented classes. We demonstrate the utility of our regularizer in learning representations for long-tailed distributions via achieving better performance than existing approaches over multiple datasets. Specifically, when applied to an unconditional GAN, it improves the FID from $13.03$ to $9.01$ on the long-tailed iNaturalist-$2019$ dataset.

\end{abstract}

\section{Introduction}
\begin{figure*}[t]
\centering
\includegraphics[width=0.98\textwidth, height=3cm]{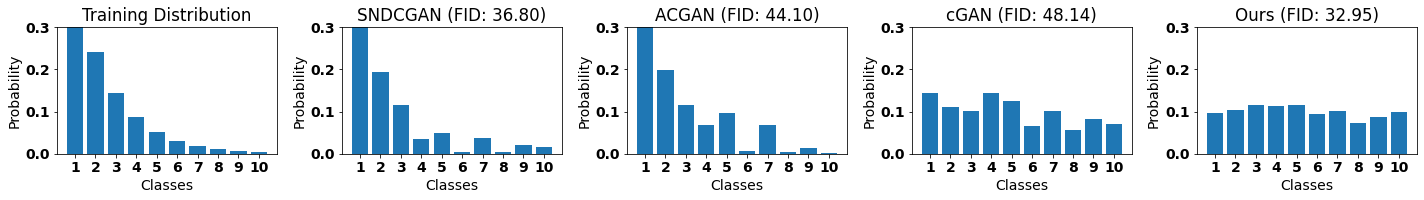}
\caption{Distribution of classes and corresponding FID scores on long-tailed CIFAR-10. SNDCGAN and ACGAN tend to produce arbitrary distributions which are biased towards majority classes, whereas cGAN samples suffer in quality with high FID. Our method achieves low FID and a balanced distribution at the same time. }
\label{fig:dist_stats}
\end{figure*}
\label{sec:introduction}
Image Generation witnessed unprecedented success in recent years following the invention of Generative Adversarial Networks (GANs) by~\citet{goodfellow2014generative}. GANs have improved significantly over time with the introduction of better architectures~\citep{gulrajani2017improved,radford2015unsupervised}, formulation of superior objective functions~\citep{jolicoeur2018relativistic,arjovsky2017wasserstein}, and regularization techniques~\citep{miyato2018spectral}. An important breakthrough for GANs has been the ability to effectively class conditioning for synthesizing images~\citep{mirza2014conditional,miyato2018cgans}. Conditional GANs have been shown to scale to large datasets such as ImageNet~\citep{imagenet_cvpr09} with $1000$ classes~\citep{miyato2018cgans}. 

One of the major issues with unconditional GANs has been their inability to produce 
balanced distributions over all the classes present in the dataset. This is seen as 
problem of missing modes in the generated distribution. A version of the
missing modes problem, known as the `covariate shift' problem was studied by~\citet{santurkar2018classification}. One possible reason is the absence of knowledge about the class distribution $P(Y|X)$\footnote{Here Y represents labels and X represents data.} of the generated samples during training. Conditional GANs on the other hand, do not suffer from this issue since the class labels $Y$ are supplied to the GAN during training. However, it has been recently found by ~\citet{ravuri2019classification} that despite being able to do well on metrics such as Inception Score (IS)~\citep{salimans2016improved} and Fr\`{e}chet Inception Distance (FID)~\citep{heusel2017gans}, the samples generated from the state-of-the-art conditional GANs lack the diversity in comparison to the underlying training datasets. Further, we observe that, although conditional GANs work well in balanced case, they suffer performance degradation in the imbalanced case (Table-\ref{tab:gan_result}). \\
In order to address these shortcomings, we propose a novel method of inducing  the information about the class distribution. We estimate the class distribution $P(Y|X)$ of generated samples in the GAN framework using a pre-trained classifier. The regularizer utilizes the estimated class distribution to penalize excessive generation of samples from the majority classes, thereby enforcing the GAN to also generate samples from minority classes. Our regularizer involves a novel method of modelling the forgetting of samples by GANs, based on the exponential forgetting observed in neural networks~\citep{kirkpatrick2017overcoming}. We show the implications of our regularizer by a theoretical upper bound in Section~\ref{sec:method}.

We also experimentally demonstrate the effectiveness of the proposed class balancing regularizer in the scenario of training GANs for image generation on long-tailed datasets, including the large scale iNaturalist-2019 \citep{inat19} dataset. Generally, even in the long-tailed distribution tasks, the test set is balanced despite the imbalance in the training set. This is because it is important to develop Machine Learning systems that generalize well across all the support regions of the data distribution, avoiding undesired over-fitting to the majority (or head) classes.  

In summary, our contributions can be listed as follows:

\begin{itemize}
\itemsep=0em
    
    \item We propose a `class-balancing' regularizer that makes use of the statistic $P(Y|X)$ of generated samples to promote uniformity while sampling from an unconditional GAN. The effect of our regularizer is depicted both theoretically (Section~\ref{sec:method}) and empirically (Section~\ref{sec:Experiments}).
    \item We show that our regularizer enables GANs to learn uniformly across classes even when the training distribution
    is long-tailed. We observe consistent gains in FID and accuracy of a
    classifier trained on the generated samples.
    
    \item Our method is able to scale to large and naturally occurring datasets such as iNaturalist-$2019$ and, achieves state-of-the-art FID score of 9.01.
\end{itemize}
\section{Background}
\label{sec:background}
\subsection{Generative Adversarial Networks (GANs)}
Generative Adversarial Network (GAN) is a two player game in which the discriminator network $D$ tries to classify images into two classes: real and fake. The generator network $G$ tries to
generate images (transforming a noise vector $z \sim P_z$ ) which fool the discriminator $D$ into classifying them as real. The game can be 
formulated as the following mathematical objective:
\begin{equation}
    \underset{G}{min} \; \underset{D}{max}\; E_{x \sim P_{r}}[\log(D(x))] + E_{z \sim P_{z}}[\log(1 - D(G(z))]
\end{equation}
The exact inner optimization for training $D$ is computationally prohibitive in large networks; hence the GAN is trained through alternate minimization of loss functions.
Multiple
loss functions have been proposed for stabilizing GAN training. In our work we use the relativistic loss function~\citep{jolicoeur2018relativistic} which is
formulated as:
\begin{equation}
    L_{D}^{rel} = - E_{(x, z) \sim (P_r, P_z)}[\log(\sigma(D(x) - D(G(z)))]
\end{equation}
\begin{equation}
L_{G}^{rel} = - E_{(x, z) \sim (P_r, P_z)}[\log(\sigma(D(G(z)) - D(x))]
\end{equation}
\textbf{Issue in the long-tailed scenario}: This unconditional GAN formulation does not use any class information $P(Y|X)$ about the images and tends to produce different number of samples from different classes~\citep{santurkar2018classification}. In other words, the generated distribution is not balanced (uniform) across different classes. This issue is more severe when the training data is long-tailed, where the GAN might completely ignore learning some (minority) classes (as shown in the SNDCGAN distribution of Figure~\ref{fig:dist_stats}).
\subsection{Conditional GAN} 
The conditional GAN~\citep{mirza2014conditional} generates images associated with a given input label
$y$ using the following objective:
\begin{equation}
    \label{equation:cgan}
    \underset{G}{\min} \; \underset{D}{\max}\; E_{x \sim P_{r}}[\log(D(x|y))] + E_{z \sim P_{z}}[\log(1 - D(G(z|y))]
\end{equation}
The Auxillary Classifier GAN (ACGAN) ~\citep{odena2017conditional} uses an auxiliary classifier for classification along with a discriminator to enforce high confidence samples from the conditioned class $y$. Whereas cGAN with projection~\citep{miyato2018cgans} uses 
Conditional Batch Norm~\citep{de2017modulating} in the generator and a projection step in the discriminator
to provide class information to the GAN. We refer to this method as cGAN in the subsequent sections.\\
\begin{figure}[!t]
    \centering\includegraphics[width=.99\linewidth]{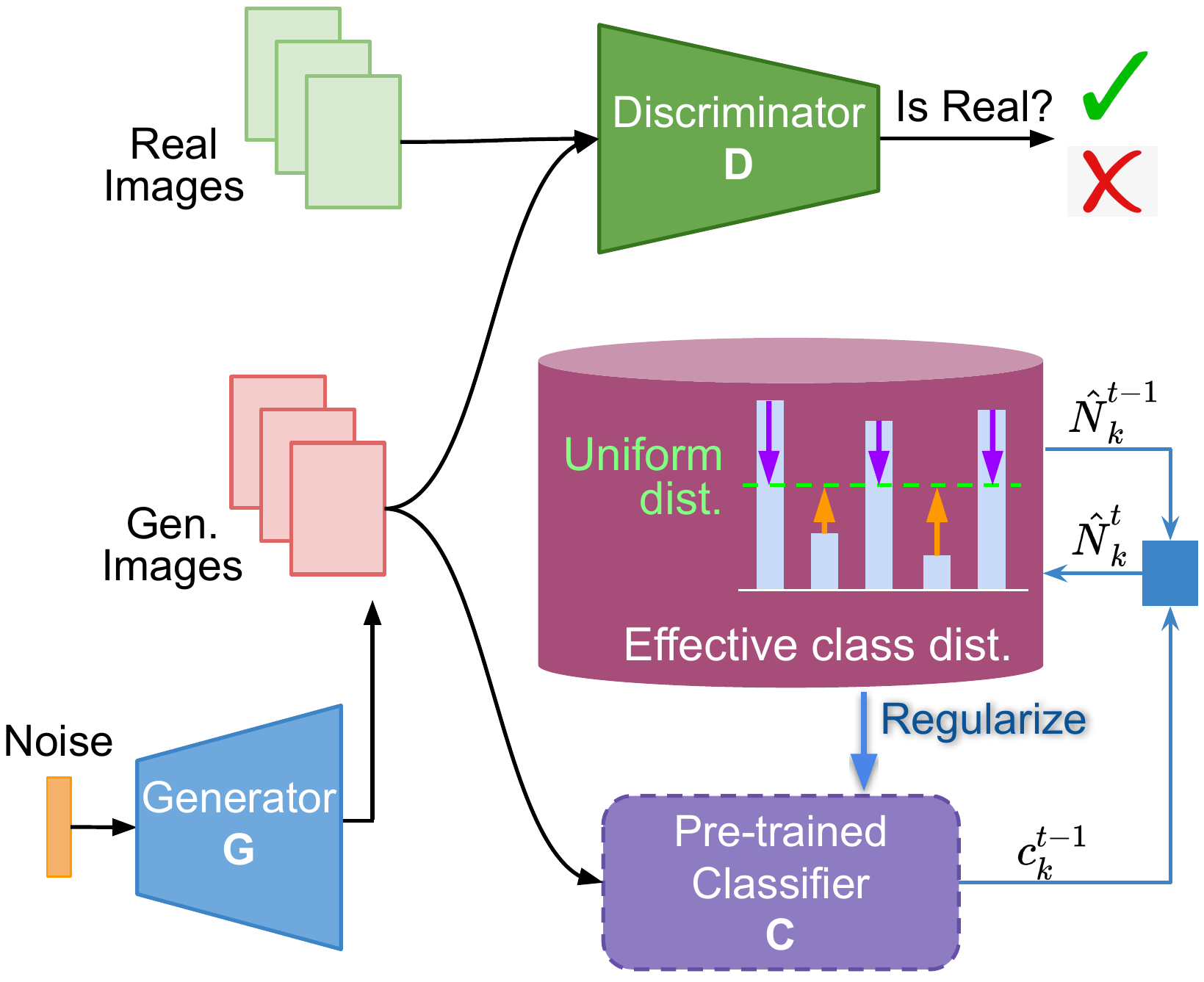}
  \caption{\textbf{Class Balancing Regularizer} aims to help GANs generate balanced distribution of samples across classes even when they are trained on imbalanced datasets. The method achieves this by keeping an estimate of effective class distribution of generated images using a pre-trained classifier. The GAN is then incentivized to generate images from underrepresented classes, which moves the GAN towards generating uniform distribution.}
  \label{fig:approach_overview}
\end{figure}

\textbf{Issue with Conditional GAN in Long-tailed Setting}:
The objective in eq.(\ref{equation:cgan}) can be seen as learning
a different $G(z|y)$ and $D(x|y)$ for each of the $K$ classes. In this
case the tail classes with fewer samples can suffer from poor learning of $G(z|y)$ as there are very few samples for learning. However, in practice there is parameter sharing among different class generators, yet there are class specific parameters present in the form of Conditional BatchNormalization. We find that performance of conditional GANs degrades more in comparison to unconditional GANs in the long-tailed scenario (observed in Table~\ref{tab:gan_result}).
\section{Method}
\label{sec:method}
In our method we use a pretrained classifier $(C)$ to provide feedback to the generator about the label distribution $P(Y|X)$ through the proposed regularizer. One can train the classifier $(C)$ before the GAN, on the underlying long-tailed dataset. However, if the full set of labels are not available, we show (Section~\ref{subsec:semi-supervised}) that a limited set of labeled data is sufficient for training the classifier. The proposed regularizer term is then added to the generator loss and trained using backpropogation. 
We first describe the method of modelling the class distribution in Section~\ref{class_statistics}. Exact formulation of the regularizer and its theoretical properties are described in Section~\ref{sub:formulation}. The overview of our method is presented in Figure~\ref{fig:approach_overview}.
\subsection{Class Statistics for GAN}
\label{class_statistics}
GAN is a dynamic system in which the generator $G$ has to continuously adapt itself in a way that it is able
to fool the discriminator $D$. During the training, discriminator $D$ updates itself, which changes the objective for the generator $G$. This change in objective can
be seen as learning of different tasks for the generator $G$. In this context, we draw motivation from the 
seminal work on catastrophic forgetting in neural networks~\citep{kirkpatrick2017overcoming} which shows that a neural network trained using SGD suffers from exponential forgetting of
earlier tasks when trained on a new task. 
Based on this, we define \emph{effective class frequency} $\hat{N_k^t}$ of class $k$ at a training cycle $t$ as:
\begin{equation}
    \hat{N_k^t} = (1 - \alpha) \hat{N_k}^{t-1} + \beta c_k^{t-1} 
    \label{eq:update_equation}
\end{equation}
Here $c_k^{t -1}$ is the number of samples of class $k$ produced by the GAN in cycle $(t-1)$ and $\hat{N_k}^0$ is initialized to a constant for all $k$ classes. The class to which the generated sample belongs to is determined by a pretrained classifier $C$. We find that using exponential decay, using either ($\beta = 1$) or a convex combination ($\beta=\alpha$) is sufficient for all the experiments. Although $D$ gets updated continuously, the update is slow and requires some iterations to change the form of $D$. Hence we update the  statistics after certain number of iterations which compose a cycle. Here $\alpha$ is the exponential forgetting factor which is set to $0.5$ as default in our experiments. We normalize the $\hat{N_k^t}$ to obtain discrete \textit{effective class distribution $N_k^t$}:
\begin{equation}
    N_k^t =  \frac{\hat{N_k^{t}}}{\sum_k \hat{N_k^{t}}}
\end{equation}

\subsection{Regularizer Formulation}
\label{sub:formulation}
The regularizer objective is defined as the minimization of the term ($L_{reg}$) below:
\begin{equation}
    \underset{\hat{p} }{\min}  \; \sum_{k} \frac{\hat{p}_k\log(\hat{p}_k)}{N_k^t}  
     \label{equation:regularizer}
\end{equation}
where $ \hat{p} = \sum_{i = 1}^{n} \frac{C(G(z_i))}{n}$ and $z_i \sim P_z$. In other words, $\hat{p}$ is the average softmax
vector (obtained from the classifier $C$) over the batch of $n$ samples and $\hat{p_k}$ is its $k^{th}$ component corresponding to class $k$. If the classifier $C$ recognizes the samples confidently with probability $\approx 1$,  $\hat{p}_k$ can be seen as the approximation to the ratio of the number of samples that belong to class $k$ to the total number of samples in the batch $n$. $N_k^t$ in the regularizer term is obtained through the update rule in Section \ref{class_statistics} and is a constant during backpropagation. The regularizer objective in Eq.~(\ref{equation:regularizer}) when multiplied with a negative unity, can also be interpreted as maximization of the weighted entropy computed over the batch.

\textbf{Proposition 1}:
The minimization of the proposed objective in (\ref{equation:regularizer}) leads to the following bound on $\hat{p_k}$: 
\begin{equation}
    \hat{p_k} \leq e^{-K(log(K) - 1)\frac{N_k^t}{\sum_{k}{N_k^t}} -1}
\end{equation}

where $K$ is the number of distinct class labels produced by the classifier $C$.

\textbf{Proof}:
\begin{equation}
    \displaystyle \underset{\hat{p}}{\min} \; \sum_{k} \frac{\hat{p_k}\log(\hat{p_k})}{N_k^t} 
\end{equation}
Introducing the probability constraint and the Lagrange multiplier $\lambda$:
\begin{equation}
    \displaystyle L(\hat{p}, \lambda) = \sum_{k} \frac{\hat{p_k}\log(\hat{p_k})}{N_k^t} - \lambda (\sum{\hat{p_k}} - 1) 
\end{equation}
On solving the equations obtained by setting $\displaystyle \frac{\partial L}{\partial \hat{p_k}} = 0:$
\begin{equation}
    \label{eq:lambda_val}
    \frac{1}{N_k^t} + \frac{\log(\hat{p_k})}{N_k^t} - \lambda = 0 \implies \hat{p_k} = e^{\lambda N_k^t - 1}
\end{equation}
Using the constraint  $\displaystyle \frac{\partial L}{\partial \lambda} = 0$ we get:
\begin{equation}
    \sum_{k} \hat{p_k} = 1 \implies \sum_{k} e^{\lambda N_k^t - 1} = 1 \implies \sum_{k} e^{\lambda N_k^t} = e
\end{equation}
Now we normalize both sides by $K$, the number of distinct labels produced by classifier and apply Jensen's inequality for concave function $\psi (\frac{\sum a_ix_i}{\sum a_i}) \geq \frac{\sum a_i\psi(x_i)}{\sum a_i}$ and take $\psi$ as $\log$ function:
\begin{equation}
\displaystyle    \frac{e}{K} = \sum_{k}\frac{e^{\lambda N_k^t}}{K} \implies \log (\frac{e}{K}) = \log(\sum_{k}\frac{e^{\lambda N_k^t}}{K}) \geq \sum_{k} \frac{\lambda N_k^t}{K}
\end{equation}
On substituting the value of $\displaystyle \lambda$ in inequality from Eq. \ref{eq:lambda_val}:
\begin{align}
    K(1 - \log(K)) & \geq \lambda \sum_{k} N_k^t \implies \\  K(1 -\log(K)) & \geq  (\sum_{k} N_k^t) \frac{1 + \log(\hat{p_k})}{N_k^t}
\end{align}
On simplifying and exponentiation we get the following result:
\begin{equation}
    \displaystyle \hat{p_k} \leq e^{-K(\log(K) - 1) \frac{N_k^t}{\sum_k N_k^t} - 1}
\end{equation}

The penalizing factor $K(\log(K) - 1)$ is increasing
in terms of number of classes $K$ in the dataset. This helps the overall objective since we need a large penalizing factor to compensate for as $N_k^t/\sum_k N_k^t$ will be smaller when number of classes is large in the dataset. Also, in case of generating a balanced distribution, $\hat{p_k} = 1/K$ which
leads to the exponential average $N_k^t = 1/K$ given sufficient iterations. In this case the upper bound value will be  $1/K$ which equals the value of $\hat{p_k}$, proving that the given bound is tight. We would like to highlight that the proof is valid for any $N_k^t > 0$ but in our case $\sum_k N_k^t = 1$. 
\\

\textbf{Implications of the proposition 1}:
  \begin{figure}[!t]
    \centering\includegraphics[width=\linewidth]{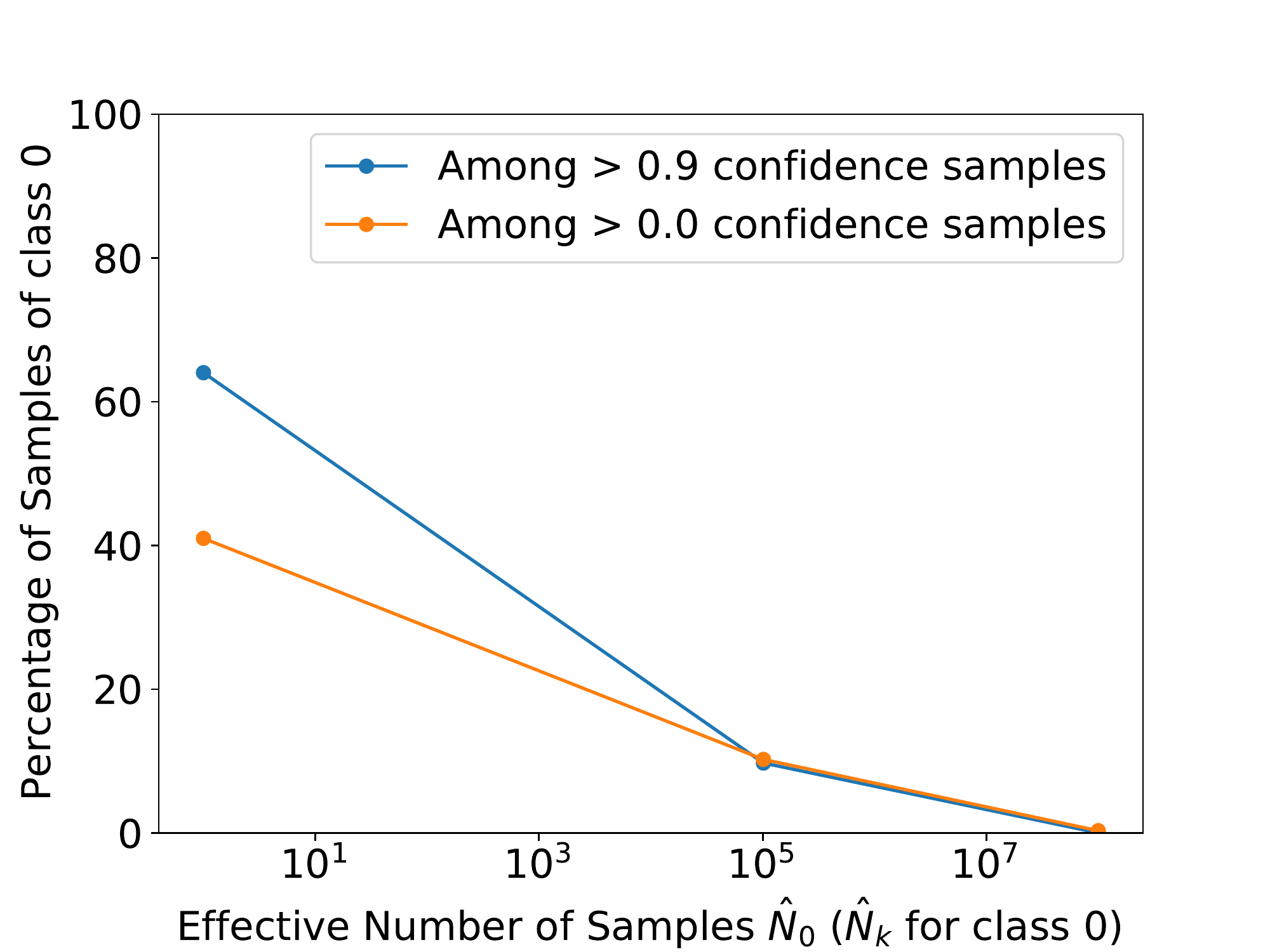}
    \caption{\label{fig:toy_exp} Shows the percentage of generated samples for class $0$ by SNDCGAN on CIFAR-10 for varying values of effective class frequency $\hat{N_0}$. When $\hat{N_0}$ is large, the network tries to decrease fraction of class 0 samples whereas when $\hat{N_0}$ is small it tries to increase fraction of class 0 samples among the generated samples. The blue and orange lines respectively correspond to the percentage of class $0$ samples, in $>0.9$ confidence samples and all the samples.}
  \end{figure}
The bound on $\hat{p_k}$ is inversely proportional to the exponent of the fraction of effective class distribution $N_k^t$ for a given class $k$. To demonstrate the effect of our regularizer empirically, we construct two extreme case
examples:

\begin{itemize}
\itemsep0em
\item If $N_k^t \gg N_i^t$, $\forall i \neq k, N_k^t \approx 1$, then the bound on $\hat{p_k}$ would approach $e^{-K(\log(K) - 1) - 1}$. Hence the network is expected to decrease the proportion of class $k$ samples.
    
\item If $N_k^t \ll N_i^t$, $\forall i \neq k, N_k^t \approx 0$, then the bound on $\hat{p_k}$ will be $e^{-1}$.  Hence the network might increase the proportion of class $k$ samples.
\end{itemize}
We verified these two extreme cases by training a SNDCGAN~\citep{miyato2018spectral} (DCGAN with spectral normalization, hyperparameters defined in Appendix Table~\ref{app: hyperparams})  on CIFAR-10 and fixing $\hat{N_k^t}$ (unnormalized version of $N_k^t$) across time steps and denote it as $\hat{N_k}$. We run two experiments by initialising $\hat{N_k}$ to a very large value and 
a very small value. Results presented in Figure~\ref{fig:toy_exp} show that the GAN increases the proportion of samples of class $k$ in case of low $\hat{N_k}$
and decreases the proportion of samples in case of large $\hat{N_k}$. This shows the balancing behaviour of proposed regularizer. \\

\textbf{Proposition 2 \citep{guiacsu1971weighted}}: If each  $\hat{p_k} = e^{\lambda N_k^t - 1}$ where $\lambda$ is obtained from solution of $\sum_{k}e^{\lambda N_k^t - 1} = 1$, then the regularizer objective in  Eq. \ref{equation:regularizer} attains the optimal minimum value of $\lambda -\sum_{k} \frac{e^{\lambda N_k^t - 1}}{N_k^t}$. For proof please refer to Appendix~\ref{app:proof2}.

\textbf{Implications of Proposition 2:} As we only use necessary conditions to prove the bound in proposition 1, the optimal solution can be a maximum, minimum or a saddle point. Prop. 2 result shows that the optimal solution found in Prop. 1 (i.e. $\hat{p_k} = e^{\lambda N_k^t - 1}$) is indeed an optimal minimum. 
\subsection{Combining the Regularizer and GAN Objective}
The regularizer can be combined with the generator loss in the following way:
\begin{equation}
    L_{g} = - E_{(x, z) \sim (P_r, P_z)}[\log(\sigma(D(G(z)) - D(x)))] + \lambda L_{reg}
\end{equation}
It has been recently shown~\citep{jolicoeur2018rfdiv} that the first term of the loss leads to minimization of $D_f(P_g, P_r)$ that is f-divergence
between real ($P_r$) and generated data distribution ($P_g$). The regularizer term ensures that the distribution of classes across generated
samples is uniform. The combined objective provides insight into the working of framework, as the first term ensures that the generated images fall in the image distribution and the second term ensures that the
distribution of classes is uniform. So the first term leads to \textit{divergence minimisation} to real data while satisfying the \textit{constraint} of class balance (i.e. second term), hence overall it can be seen as \textit{constained optimization}.

As $P_r$ comprises of diverse samples from majority class the first objective
term ensures that $P_g$ is similarly diverse. The second term in the objective ensures that the discriminative properties
of all classes are present uniformly in the generated distribution, which ensures that 
minority classes get benefit of diversity present in the majority classes. This is analogous to approaches that transfer knowledge from
majority to minority classes for long-tailed classifier learning~\citep{Liu_2019_CVPR,wang2017learning}.

\begin{table*}[ht]
    \centering
    \begin{tabular}{lccccccc}
    \hline
    Imbalance Ratio &  \multicolumn{3}{c}{100} & \multicolumn{3}{c}{10} &  \multicolumn{1}{c}{1}\\\hline
                & FID ($\downarrow$)&      KLDiv($\downarrow$)&  Acc.($\uparrow$)& FID($\downarrow$)&      KLDiv($\downarrow$)&  Acc.($\uparrow$)& FID ($\downarrow$)    \\\hline 
   \multicolumn{8}{c}{CIFAR-10} \\ \hline
         SNDCGAN	&36.97 $\pm$ 0.20&	0.31 $\pm$ 0.0& 68.60 & 32.53 $\pm$ 0.06& 	0.14 $\pm$ 0.0 &  80.60&	27.03 $\pm$ 0.12\\
         ACGAN	&44.10 $\pm$ 0.02 &	0.33 $\pm$ 0.0&	43.08& 38.33 $\pm$ 0.10&	0.12 $\pm$ 0.0&	60.01& 24.21 $\pm$ 0.08 \\
         cGAN	& 48.13 $\pm$ 0.01&	0.02 $\pm$ 0.0&	47.92& 26.09 $\pm$ 0.04&	0.01 $\pm$ 0.0&	68.34& 18.99 $\pm$ 0.03\\
         Ours	&32.93 $\pm$ 0.11 &	0.06 $\pm$ 0.0&	72.96& 30.48 $\pm$ 0.07&	0.01 $\pm$ 0.0& 82.21& 25.68$\pm$ 0.07 \\
         \hline 
        \multicolumn{8}{c}{LSUN} \\ \hline    
         SNResGAN	&37.70 $\pm$ 0.10 &	0.68 $\pm$ 0.0 &	75.27& 33.28 $\pm$ 0.02 & 	0.29 $\pm$ 0.0& 79.20&	28.99 $\pm$ 0.03 \\
         ACGAN	&43.76 $\pm$ 0.06&	0.39 $\pm$ 0.0 &	62.33& 31.98 $\pm$ 0.02&	0.05 $\pm$ 0.0&	75.47& 26.43 $\pm$ 0.04\\
         cGAN	&75.39 $\pm$ 0.12&	0.01 $\pm$ 0.0   & 	44.40 & 30.68 $\pm$ 0.04&	0.00 $\pm$ 0.0&	72.93 & 27.59 $\pm$  0.03 \\
         Ours	&35.04 $\pm$ 0.19&	0.06 $\pm$ 0.0&	77.93& 28.78 $\pm$ 0.01&	0.01 $\pm$ 0.0&	82.13& 28.15 $\pm$ 0.05\\
         \hline

    \end{tabular}
    \caption{Results on CIFAR-10 (top panel) and $5$ class subset of LSUN (bottom panel) datasets with varying imbalance. In the last column FID values in balanced scenarios are present for
    ease of reference. FID, KL Div. and Acc. are calculated on $50$K sampled images from each GAN.}
    \label{tab:gan_result}
\end{table*}

\label{formulation}
\section{Experiments}
\label{sec:Experiments}
For evaluating the effectiveness of our balancing regularizer, we conduct image generation experiments over long-tailed distributions. 
In these experiments we aim to train a GAN using data from a long-tailed dataset, which is common in the real world setting. For achieving good performance on all classes in the dataset, the method requires to transfer knowledge from majority to minority classes. Several works have focused on learning classifiers on long-tailed distributions~\citep{cao2019learning,cui2019classbalancedloss, Liu_2019_CVPR}. Yet, works focusing on Image Generation using long-tailed dataset are limited. Generative Minority Oversampling (GAMO)~\citep{Mullick_2019_ICCV} attempts to solve the problem
by introducing a three player framework, which is an encoder-decoder network and not a GAN. We do not compare our results
with GAMO as it is not 
trivial to extend GAMO to use schemes such as Spectral Normalization \citep{miyato2018spectral}, and ResGAN like architecture~\citep{gulrajani2017improved} which impede fair comparison.

\textbf{Datasets}: We perform extensive experimentation on CIFAR-10 and a subset of LSUN, as these are widely used for evaluating GANs. The LSUN subset consists of $250$K training images and $1.5$K validation images. 
The LSUN subset is composed of $5$ balanced classes. \citet{santurkar2018classification} identified this subset to be a challenging case for GANs to generate uniform distribution of classes.
The original CIFAR-10 dataset is composed of $50$K training images and $10$K validation images. We construct the
long-tailed version of the datasets by following the same 
procedure as~\citet{cao2019learning}.
Here, images are removed from training set to convert it
to a long-tailed distribution while the validation set is kept unchanged.
The imbalance ratio $(\rho)$ determines the ratio of number of samples in most populated
class to the least populated one: $\displaystyle \rho = {max_{k}\{n_k\}}/{min_{k}\{n_k\}}$.

\textbf{Pre-Trained Classifier}:
\label{sec:pretrained_classifier}
An important component of our framework is the pre-trained classifier. All the pre-trained classifiers in our experiments use a ResNet32~\citep{he2016deep} architecture.
The classifier is trained using Deferred Re-Weighting (DRW) scheme~\citep{cao2019learning,cui2019classbalancedloss} on the long-tailed data. We use the available open source code\footnote{https://github.com/kaidic/LDAM-DRW}. We use the following learning 
rate schedule: initial learning rate of $0.01$ and multiplying by
$0.01$ at epoch $160$ and $180$. We train the models for 
$200$ epochs and start re-weighting (DRW) at epoch $160$. We give a summary of the validation accuracy of the models for various imbalance ratios $(\rho)$ in Table~\ref{tab:clf_acc}.

\begin{table}[h]
    \centering
    \begin{tabular}{lccc}
         \hline
         Imbalance Ratio $(\rho)$ & 100 & 10 & 1  \\ \hline
         CIFAR-10 & 76.67 & 87.70 & 92.29  \\ 
         LSUN & 82.40 & 88.07 &  90.53\\ \hline
    \end{tabular}
    \caption{Validation Accuracy of the PreTrained Classifiers used with GANs. The balanced classifier also serves as an annotator.}
    \label{tab:clf_acc}
\end{table}

\textbf{GAN Architecture}: 
We use the SNDCGAN architecture for experiments on CIFAR-10 with images of size of $32 \times 32$ and SNResGAN (ResNet architecture with spectral normalization) structure for experiments on LSUN dataset with images of size $64 \times 64$. For the conditional GAN baselines we conditioned the generator using Conditional BatchNorm.
We compare our method to two widely used conditional GANs: ACGAN and cGAN. The other baseline we use is the unconditional GAN (SNDCGAN \& SNResGAN) without our regularizer. All the GANs were trained with spectral normalization in the discriminator for stabilization~\citep{miyato2018spectral}.

\textbf{Training Setup:}
We use the learning rate of $0.0002$
for both generator and discriminator. We use Adam optimizer with $\beta_1 = 0.5$
and $\beta_2 = 0.999$ for SNDCGAN and $\beta_1 = 0$ and $\beta_2 = 0.999$
for SNResGAN. We use a batch size of $256$ and perform
$1$ discriminator update for every generator update. As a sanity check, we use
the FID values and visual inspection of images on the balanced dataset and verify the range of values from~\citet{kurach2019large}. We update the statistics $N_k^t$ using Eq. \ref{eq:update_equation} after every 2000 iterations, for all experiments in Table \ref{tab:gan_result}. The code is implemented with PyTorch Studio~\cite{kang2020contrastive} GAN framework. Further details and ablations are present in the Appendix.

\textbf{Evaluation} We used the following evaluation metrics:

\textbf{1. KL Divergence w.r.t. Uniform Distribution of labels}: Labels for the generated samples are obtained by using the pre-trained classifier (trained on balanced data) as an annotator. The annotator is just used for evaluation on long-tailed data. Low values of this metric signify that the generated samples are uniformly distributed across classes. \\
\textbf{2. Classification Accuracy (CA)}: We use the $\{(X, Y)\}$ pairs from the GAN generated samples to train a ResNet32 classifier and test it on real data. For unconditional GANs, the label $Y$ is obtained from the classifier trained on long-tailed data. Note that this is similar to Classifier Accuracy Score of \citep{ravuri2019classification}.  \\
\textbf{3. Fr\`{e}chet Inception Distance (FID)}: It measures the 2-Wasserstein Distance on distributions obtained from Inception Network~\citep{heusel2017gans}. We use $10$K samples from CIFAR-10 validation set and $10$K ($2$K from each class) fixed random images from
LSUN dataset for measuring FID.\\

\textbf{Discussion of Results:} We present our results below: \\
1) \textbf{Stability}: In terms of stability, we find that cGAN suffers from 
early collapse in case of high imbalance ($\rho = 100$) and stops improving within $10$K iterations. Therefore, although cGAN is stable in balanced scenario, it
is unstable in case of long-tailed version of the given dataset.\\
2) \textbf{Biased Distribution}: Contrary to cGAN, we find that the distribution of classes generated by ACGAN, SNDCGAN and SNResGAN is imbalanced. The images obtained by sampling uniformly and labelling by annotator, suffers from a high KL 
divergence to the uniform distribution. Some classes are almost absent from the distribution of generated samples as shown in Figure \ref{fig:dist_stats}. In this case, in Table~\ref{tab:gan_result} we observe FID score
just differs by a small margin even if there is a large imbalance in class distribution. This happens for ACGAN as its loss is composed of GAN loss and classification loss terms, therefore in the long-tailed setting the ACGAN gets biased towards optimising GAN loss ignoring classification loss, hence tends to produce arbitrary distribution.  Contrary to this, our GAN produces class samples uniformly as it is evident from the low KL Divergence.\\
3) \textbf{Comparison with State-of-the-Art Methods}: In this work we observe that classification accuracy is weakly correlated with FID score
which is in agreement with~\citet{ravuri2019classification}. We achieve better classifier accuracy when compared to
cGAN in all cases, which is the current state-of-the-art for Classifier Accuracy Score (CAS). Our method shows minimal degradation in FID in comparison to the corresponding balanced case. It is also able to achieve the best FID in $3$ out of $4$ long-tailed cases. Hence we expect that methods such as Consistency Regularization~\citep{zhang2019consistency} and Latent Optimization~\citep{wu2019logan} can be applied in conjunction with our method to further improve
the quality of images. However in this work we specifically focused
on techniques used to provide class information $(Y)$ of the images $(X)$ to the GAN. Several state-of-the-art GANs use the approach of cGAN~\citep{wu2019logan,brock2018large} for conditioning the discriminator
and the generator.

\subsection{Experiments on Datasets with Large Number of Classes}
For showing the effectiveness of our method on datasets with large number of classes, we show results on long-tailed CIFAR-100 ($\rho$ = 10) and iNaturalist-2019~\citep{inat19} (1010 classes). The iNaturalist dataset consists of image of species which naturally have a long-tailed distribution. For the iNaturalist dataset we use a batch size of 256 which is significantly less than number of classes present (1010), and use $\beta = \alpha$ for updating the \textit{effective class distribution} in Eq.~\ref{eq:update_equation}. We use a SNResGAN based architecture for both datasets.  Additional hyper-parameters and training details are present in Appendix Section \ref{app:largedatasets}. We use CIFAR-100 validation set and a balanced subset of (16160) iNaturalist images for calculation of FID with $50$K generated images in each case.

\begin{figure*}
     \centering
     \begin{subfigure}[b]{0.22\textwidth}
         \centering
         \includegraphics[width=\textwidth]{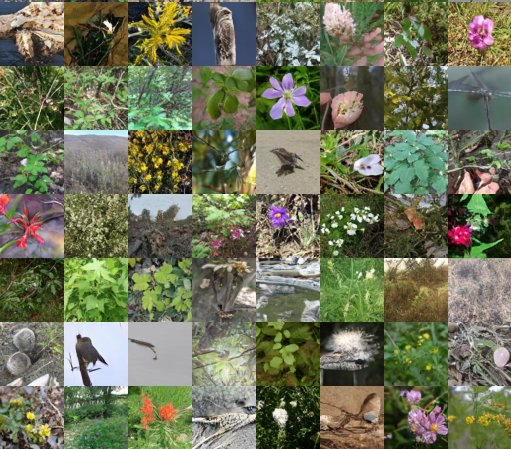}
         \caption{SNResGAN (13.03 FID)}
         \label{fig:y equals x}
     \end{subfigure}
     \hfill
     \begin{subfigure}[b]{0.22\textwidth}
         \centering
         \includegraphics[width=\textwidth]{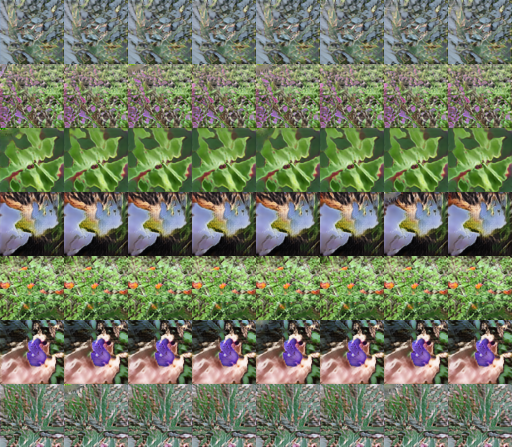}
         \caption{ACGAN (47.15 FID)}
         \label{fig:three sin x}
     \end{subfigure}
     \hfill
     \begin{subfigure}[b]{0.22\textwidth}
         \centering
         \includegraphics[width=\textwidth]{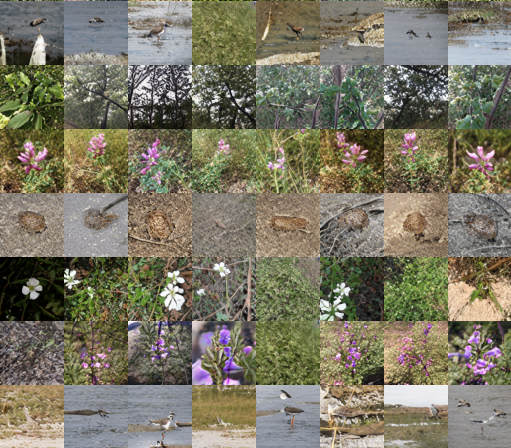}
         \caption{cGAN (21.53 FID)}
         \label{fig:five over x}
     \end{subfigure}
     \hfill
      \begin{subfigure}[b]{0.22\textwidth}
         \centering
         \includegraphics[width=\textwidth]{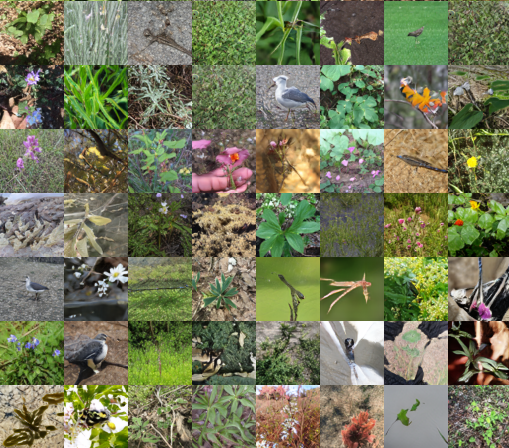}
         \caption{Ours (9.01 FID)}
         \label{fig:y equals x}
     \end{subfigure}
        \caption{Shows the $64 \times 64$ generated images for the iNaturalist-2019 dataset for different baselines.}
        \label{fig:inat_visual}
    
\end{figure*}

Table \ref{tab:cifarinat} summarizes the results on the above two datasets. Our method clearly outperforms all the other baselines in terms of FID. For the CIFAR-100 dataset our method achieves balanced distribution similar to the cGAN. In the case of iNaturalist our method achieves KL DIV of $0.6$ which is significantly better than other baselines. Other baselines cause significant imbalance in the generated distribution and hence are unsuitable for real world long-tailed datasets such as iNaturalist (examples are shown in Figure \ref{fig:inat_visual}. The superior results on iNaturalist demonstrate that our method is also effective  in the case when batch size is less than the number of classes present in the dataset. 

\begin{table}[t]
    \centering
    \resizebox{\columnwidth}{!}{%
    \begin{tabular}{lcccc}
    \hline
    & FID ($\downarrow$)&      KLDiv($\downarrow$) & FID ($\downarrow$)&      KLDiv($\downarrow$) \\ \hline
    Imbalance Ratio & \multicolumn{2}{c}{100} & \multicolumn{2}{c}{10} \\ \hline
    \multicolumn{5}{c}{CIFAR-10} \\ \hline 
        \pbox[c][0.5 cm] [c] {10cm}{SNDCGAN} & 36.97 $\pm$ 0.20&	0.31 $\pm$ 0.0&  32.53 $\pm$ 0.06& 	0.14 $\pm$ 0.0 \\ \hline
        \pbox[c][1. cm] [c] {10cm}{Ours\\ (Supervised)}  	&32.93 $\pm$ 0.11 &	0.06 $\pm$ 0.0& 30.48 $\pm$ 0.07&	0.01 $\pm$ 0.0\\ \hline
        \pbox[c][1. cm] [c] {3cm}{Ours\\ (Semi Supervised)} & 33.32 $\pm$ 0.03 & 0.14 $\pm$ 0.0 &  30.37 $\pm$ 0.14 &  0.04 $\pm$ 0.0 \\ \hline
        \multicolumn{5}{c}{LSUN} \\ \hline 
        
        \pbox[c][0.5 cm] [c] {10cm}{SNResGAN} & 37.70 $\pm$ 0.10 &	0.68 $\pm$ 0.0 &   33.28 $\pm$ 0.02 & 	0.29 $\pm$ 0.0 \\ \hline
        \pbox[c][1. cm] [c] {10cm}{Ours\\ (Supervised)} &  35.04 $\pm$ 0.19&	0.06 $\pm$ 0.0& 28.78 $\pm$ 0.01&	0.01 $\pm$ 0.0  \\ \hline
        \pbox[c][1. cm] [c] {10cm}{Ours\\ (Semi Supervised)}  & 35.95 $\pm$ 0.05 & 0.15 $\pm$ 0.0 & 30.96 $\pm$ 0.07 & 0.06 $\pm$ 0.0 \\ \hline
    \end{tabular}}
    \caption{Comparison of results in Semi Supervised Setting. The pretrained classifier used in our framework is fine-tuned with 0.1\% of labelled data. The same classifier trained on balanced dataset is used as annotator for calculating KL Divergence for all baselines.}
    \label{tab:semi-sup}
    
\end{table}
\subsection{Other baselines using Pre-trained Classifier}
Since the proposed generative framework utilizes a pretrained classifier, we believe the comparison (against other generative model) should be performed via providing the same classifier. Therefore, in this subsection, we choose ACGAN framework and build a baseline by adding the pretrained classifier. Note that in ACGAN discriminator not only performs the real-fake distinction but also serves as an auxiliary classifier and labels the sample into the underlying classes in the dataset. In this baseline, we replace the latter part with the pre-trained classifier used in the proposed framework. Hence both the frameworks are even with respect to the availability of the $P(Y/X)$ information. 

\begin{figure*}[h]
    \centering
    \includegraphics[width=0.75\textwidth]{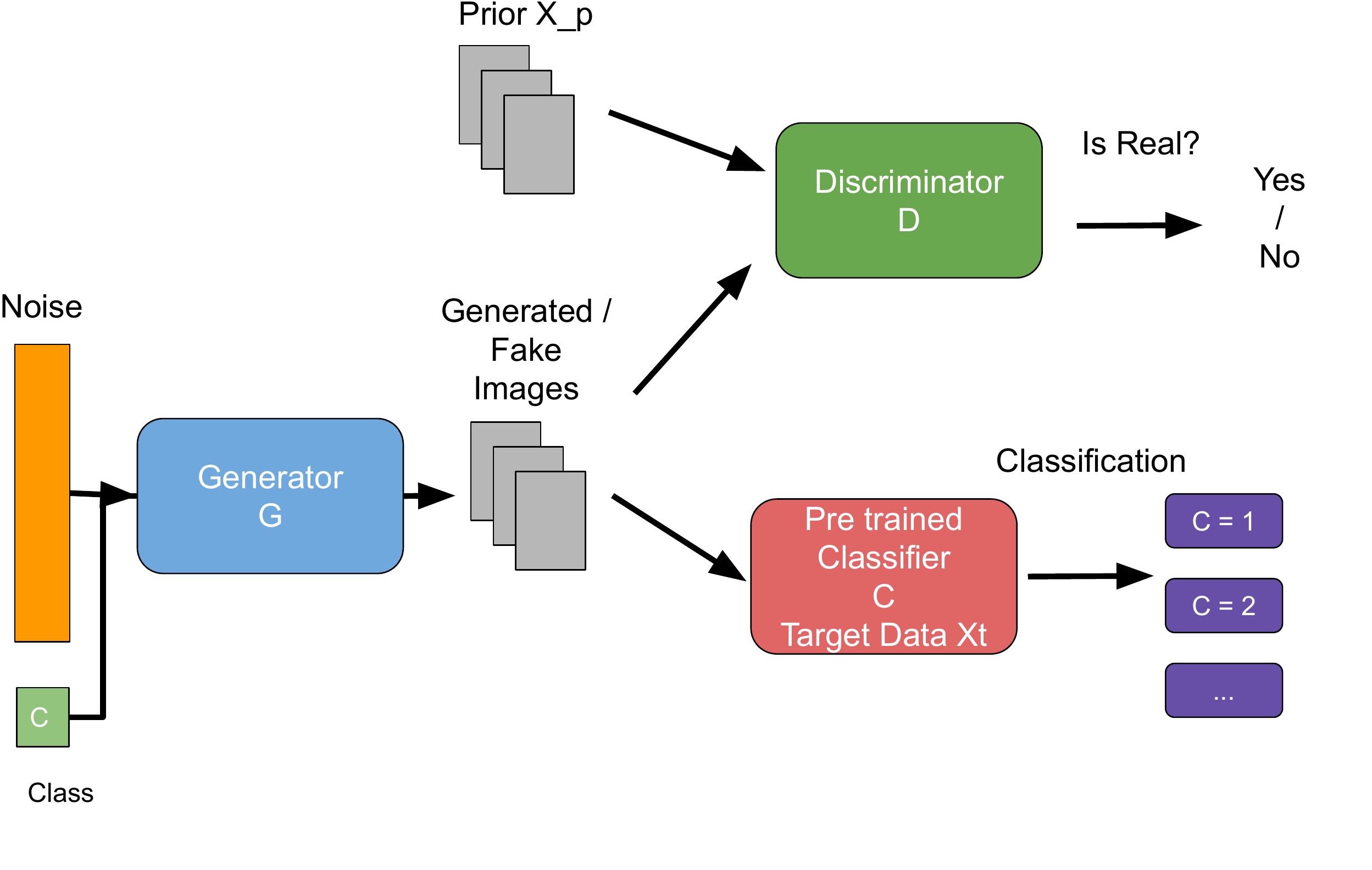}
    \caption{Diagram showing the overview of the other baseline.}
    \label{fig:approach_other_baseline}
\end{figure*}

The resulting generator can avail the label information of the generated samples from this classifier. In other words, if the generator intends to produce a sample of class $y$ via conditioning, the pretrained classifier can provide the required feedback in the form of cross entropy loss. However, we find that this baseline of ACGAN that employs a pre-trained classifier suffers from mode collapse ($42.28$ FID) and only generates extremely limited within class diversity in images (e.g. in Fig. \ref{fig:other_baseline_images}). On the contrary, our method ($9.01$ FID) using the same pre-trained classifier doesn't suffer from mode collapse and also works in case of iNaturalist dataset. This shows that the proposed framework and regularizer are non-trivial and prevent the GAN from mode collapse. We believe there is scope for understanding the nature of the involved optimization in this future. For the exact details of implementation please refer to the Appendix Section~\ref{app:other_baseline}. An overview of the baseline is depicted in Figure \ref{fig:approach_other_baseline}.
\subsection{Semi-supervised class-balancing GAN}
\label{subsec:semi-supervised}
In case of conditional GANs, class labels are required for GAN training. However, in our case the stage of classifier learning which requires labels is decoupled from GAN learning. In this section we show how this can be advantageous in practice. Since our framework only requires knowledge of $P(Y/X)$, we find that a classifier trained through any of a variety of sources could be used for providing feedback to the Generator. This feedback allows the GAN to generate class balanced distributions even in cases when the labels for underlying long-tailed distributions are not known. This reduces the need for labelled data in our framework and shows the effectiveness over conditional GAN. Note that the performance of conditional GANs deteriorates~\citep{lucic2019high} when used with limited labelled data. We use a ResNet-50 pretrained model on ImageNet from BiT (Big Image Transfer)~\citep{kolesnikov2019large} and finetune it using $0.1$ \% of labelled data of balanced training set (i.e. $5$ images per class for CIFAR-10 and $50$ images per class for LSUN dataset). 

\begin{table}[t]
    \centering
    \resizebox{\columnwidth}{!}{%

    \begin{tabular}{lcc|cc}
        \hline
         & \multicolumn{2}{c}{iNatuarlist 2019} & \multicolumn{2}{c}{CIFAR-100} \\\hline
         & FID ($\downarrow$) & KL Div($\downarrow$)          & FID ($\downarrow$) & KL Div($\downarrow$)  \\ \hline
         SNResGAN & 13.03 $\pm$ 0.07 & 1.33 $\pm$ 0.0 & 30.05 $\pm$ 0.05  & 0.18 $\pm$ 0.0\\  
         ACGAN & 47.15 $\pm$ 0.11 & 1.80 $\pm$ 0.0 & 69.90 $\pm$ 0.13 & 0.40 $\pm$ 0.0\\ 
         cGAN & 21.53 $\pm$ 0.14 & 1.47 $\pm$ 0.0 & 30.87 $\pm$ 0.06 &  0.09 $\pm$ 0.0 \\
         Ours & 9.01 $\pm$ 0.08 & 0.60 $\pm$ 0.0 & 28.17 $\pm$ 0.06 & 0.11 $\pm$ 0.0 \\
         \hline

    \end{tabular}}
    \caption{Results on iNaturalist (2019) and CIFAR-100 ($\rho=10$) dataset. Significant performance increase is achieved by our method in comparison to baselines.}
    \label{tab:cifarinat}
\end{table}
\vspace{-3.0mm}
\subsection{Analysis on the Effect of Classifier Performance}
\begin{figure}[!ht]
    \centering
    \includegraphics[width=\columnwidth]{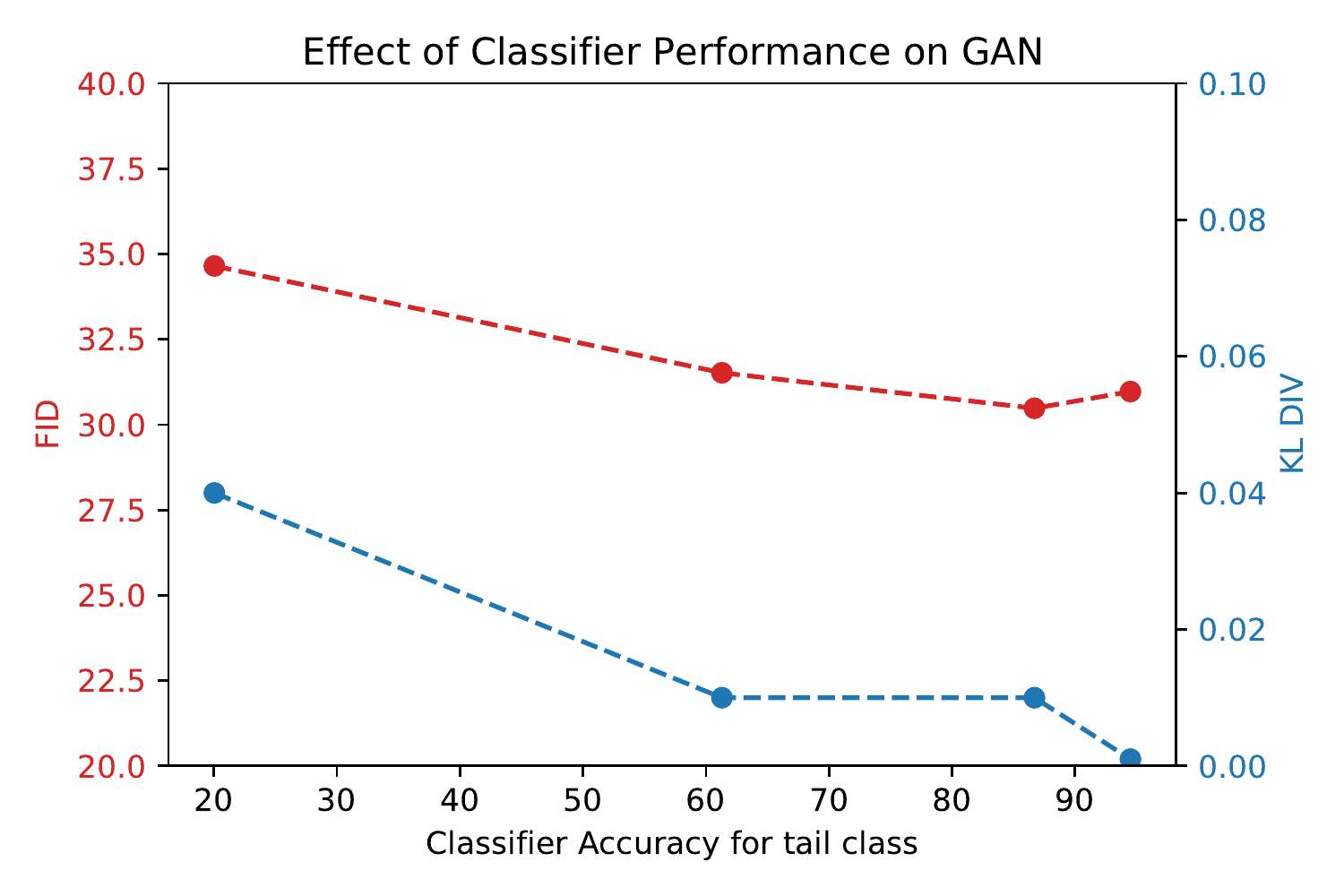}
    \caption{Analysis of Classifier Performance on long-tailed CIFAR-10 ($\rho$ = 10). The data points from left to right correspond to imbalance ratio's ($\rho$) 500, 100, 10 and 1 respectively. The performance is significantly robust after tail class accuracy reaches 60\%.}
    \label{fig:clf_analysis}
\end{figure}
For analyzing the effect of classifier performance on the GAN training in our method, we train the GAN on long-tailed CIFAR-10 ($\rho = 10$) with different classifiers. We learn multiple classifiers with different imbalance ratios ($\rho$) of $1, 10, 100, 500$ and $5000$. As the imbalance ratio increases, the accuracy of the resulting classifier decreases, hence, we can have classifiers of varied accuracy for deploying in our framework. Note that in case of high imbalance ratio, it becomes harder for the classifier to learn the tail part of the distribution. Figure~\ref{fig:clf_analysis} shows the performance (FID and KLDIV) of the resulting GAN with respect to the classifier performance on the tail class (i.e. least populated class).

From Figure~\ref{fig:clf_analysis} it can be observed that for a large range of classifier accuracies on the tail class, GANs learned in our framework are able to achieve similar FID and KL Divergence performance. Our framework only requires reasonable classifier performance which can be easily achieved via normal cross entropy training on the long-tailed dataset. For further enforcing this claim, we only use a classifier that is trained using cross entropy loss for iNaturalist-2019 experiments, in which our method achieves state-of-the-art FID score. Hence one does not need to explicitly resort to any complex training regimes for the classifier. In the extreme case of $\rho = 5000$, the classifier performance is $0$ on the tail class. In this case, our algorithm diverges and is not able to produce meaningful results. 
However, note that with a classifier accuracy of as small as $20\%$ our framework achieves decent FID and KL Divergence.
\vspace{-0.3cm}
\section{Conclusion}
\label{section:conclusion}
Long-tailed distributions are a common occurrence in real-world datasets in the context of learning problems such as object recognition. However, a vast majority of the existing contributions consider simplified laboratory scenarios in which the data distributions are assumed to be uniform over classes. In this paper, we consider learning a generative model (GAN) on long-tailed data distributions with an objective to faithfully represent the tail classes. We propose a class-balancing regularizer to balance class distribution of generated samples while training the GAN. We justify our claims on the proposed regularizer by presenting a theoretical bound and comprehensive experimental analysis. The key idea of our framework is having a classifier in the loop for keeping an uninterrupted check on the GAN's learning which enables it to retain the minority nodes of the underlying data distribution. We demonstrated that the dependency on such a classifier is not arduous. Our experimental analysis clearly brings out the effectiveness of our regularizer in the GAN framework for generating the images from complex long-tailed datasets such as iNaturalist, on which it achieves the state-of-the-art performance.
\vspace{-0.5cm}
\begin{acknowledgements} 
    Harsh Rangwani acknowledges the support from Prime Minister's Research Fellowship (PMRF). This work was supported by SERB, DST, Govt. of India (Project: STR/2020/000128). We thank Sravanti Addepalli, Gaurang Sriramanan and other Video Analytics Lab members for their valuable feedback.
\end{acknowledgements}

\bibliography{paper}
\clearpage

\providecommand{\upGamma}{\Gamma}
\providecommand{\uppi}{\pi}

\section{Appendix}
\label{appendix}
\subsection{Proof of Proposition 2}
\label{app:proof2}
\textbf{Proposition 2 \citep{guiacsu1971weighted}:} If each $\hat{p_k}$ satisfies Eq. \ref{eq:lambda_val}, i.e. $\hat{p_k} = e^{\lambda N_k^t - 1}$, then the regularizer objective in  Eq. \ref{equation:regularizer} attains the optimal minimum value of $\lambda -\sum_{k} \frac{e^{\lambda N_k^t - 1}}{N_k^t}$. \\
\textbf{Proof:} We wish to optimize the following objective:
\begin{equation}
    \displaystyle \underset{\hat{p}}{\min} \; \sum_{k} \frac{\hat{p_k}\log(\hat{p_k})}{N_k^t} \quad \text{such that} \quad \sum_{k} \hat{p_k} =1
\end{equation}
Introducing the probability constraint via the Lagrange multiplier $\lambda$:
\begin{equation}
    \displaystyle L(\hat{p}, \lambda) = \sum_{k} \frac{\hat{p_k}\log(\hat{p_k})}{N_k^t} - \lambda (\sum{\hat{p_k}} - 1) 
\end{equation}
\begin{equation}
    \displaystyle L(\hat{p}, \lambda) - \lambda = \sum_{k} \frac{\hat{p_k}\log(\hat{p_k})}{N_k^t} - \lambda (\sum{\hat{p_k}}) 
\end{equation}
\begin{equation}
    \displaystyle L(\hat{p}, \lambda) - \lambda = \sum_{k} \frac{\hat{p_k}\log(\hat{p_k}e^{-\lambda N_k^t})  }{N_k^t} 
\end{equation}
\begin{equation}
    \displaystyle L(\hat{p}, \lambda) - \lambda = \sum_{k} \frac{\hat{p_k}\log(\hat{p_k}e^{-\lambda N_k^t})  }{N_k^t} 
\end{equation}
\begin{equation}
    \displaystyle L(\hat{p}, \lambda) - \lambda = \sum_{k} \frac{e^{\lambda N_k^t}}{N_k^t} ({\hat{p_k}e^{-\lambda N_k^t}\log(\hat{p_k}e^{-\lambda N_k^t}))  }
\end{equation}
Now, for $x>0$, we know that $x\log(x) \geq -\frac{1}{e}$:
\begin{equation}
    \displaystyle L(\hat{p}, \lambda)  \geq \lambda -\sum_{k} \frac{e^{\lambda N_k^t - 1}}{N_k^t}
\end{equation}
The $xlog(x)$ attains the minimal value at only $x = \frac{1}{e}$, which is the point corresponding to $\hat{p_k}e^{-\lambda N_k^t} = 1/e$, for every $k$. This shows that at $\hat{p_k} = e^{\lambda N_k^t - 1}$ is the optimal point where the objective attains it's minimum value. This result is derived from Theorem 2 in \citet{guiacsu1971weighted}.

\subsection{Datasets}
\label{app: datasets}
We use CIFAR-10 \citep{krizhevsky2009learning} dataset for our experiments which has $50$K training
images and $10$K validation images. For the LSUN \citep{journals/corr/YuZSSX15} dataset we use a fixed
subset of $50$K training images for each of bedroom, conference room, dining room, kitchen and living room classes. In total we have $250$K training images and $1.5$K validation set of images for LSUN dataset. The imbalanced versions of the datasets are created by removing images from the training set. For the large dataset experiments, we make use of CIFAR-100 \citep{Krizhevsky09learningmultiple} and iNaturalist-2019 \citep{inat19}. The CIFAR-100 dataset is composed of the 500 training images and 100 testing images for each class. The iNaturalist-2019 is a long-tailed dataset composed of the 268,243 images present across 1010 classes in the training set, the validation set is composed of 3030 images balanced across classes.

\subsection{Architecture Details for GAN}
\label{apx:gan_arch}
We use the SNDCGAN architecture for experiments on CIFAR-10 and SNResGAN architecture for experiments
on LSUN, CIFAR-100 and iNaturalist-2019 datasets \citep{gulrajani2017improved, miyato2018spectral}. The notation for the architecture tables
are as follows: m is the batch size, FC(dim\_in, dim\_out) is a fully connected Layer, CONV(channels\_in, channels\_out, kernel\_size, stride) is convolution layer, TCONV(channels\_in, channel\_out, kernel\_size, stride) is the transpose convolution layer, BN is BatchNorm \citep{ioffe2015batch} Layer in case of unconditonal
GANs and conditional BatchNorm in case of conditional GANs. LRelu is the leaky relu activation function and
GSP is the Global Sum Pooling Layer. The DIS\_BLOCK(channels\_in, channels\_out, downsampling) and GEN\_BLOCK(channels\_in, channels\_out, upsampling) correspond to the Discriminator and
Generator block used in \citet{gulrajani2017improved}. The SNResGAN architecture for CIFAR-100 differs by a small amount as it has $32 \times 32$ image size, for which we use the exact same architecture described in \citet{miyato2018spectral}. The architectures are presented in detail
in Tables \ref{SNDCGAN_G}, \ref{SNDCGAN_D}, \ref{SNResGAN_G}  and \ref{SNResGAN_D}.
\begin{table}[h]
    \centering
    \begin{tabular}{lccc}
        \hline
        Imbalance Ratio ($\rho$) & 100  &  10  & 1\\ \hline
        CIFAR-10 & 10 & 7.5 & 5\\ 
        LSUN & 20 & 7.5 & 5\\ \hline
        
    \end{tabular}
    \caption{Values of $\lambda$ for different imbalance cases. For LSUN the $\lambda$ gets divide by 5 and for $\lambda$ it gets divided by 10 before multiplication to regularizer term.}
    \label{tab:my_label}
\end{table}

\begin{table*}[p]
  
\centering
  \begin{tabular}{llrc}
    \toprule
    \textbf{Layer} & \textbf{Input}&\textbf{Output} & \textbf{Operation}\\
    \midrule
    Input Layer & (m, 128)&(m, 8192)&\textsc{FC}(128, 8192)\\
    \midrule
    Reshape Layer & (m, 8192)&(m, 4, 4, 512)&\textsc{Reshape}\\
    Hidden Layer & (m, 4, 4, 512)&(m, 8, 8, 256)&\textsc{TConv}(512, 256, 4, 2),\textsc{BN},\textsc{LRelu} \\
    Hidden Layer & (m, 8, 8, 256)&(m, 16, 16, 128)&\textsc{TConv}(256, 128, 4, 2),\textsc{BN},\textsc{LRelu} \\
    Hidden Layer & (m, 16, 16, 128)&(m, 32, 32, 64)&\textsc{TConv}(128, 64, 4, 2),\textsc{BN},\textsc{LRelu} \\
    Hidden Layer & (m, 32, 32, 64)&(m, 32, 32, 3)&\textsc{Conv}(64, 3, 3, 1) \\
    \midrule
    Output Layer & (m, 32, 32, 3)&(m, 32, 32, 3)&\textsc{Tanh} \\
    \bottomrule
  \end{tabular}
   \caption{Generator of SNDCGAN~\citep{miyato2018spectral, radford2015unsupervised} used for CIFAR10 image synthesis.}
   \label{SNDCGAN_G}
\end{table*}
 
\begin{table*}[p]

  \centering
  \begin{tabular}{llrc}
    \toprule
    \textbf{Layer} & \textbf{Input} & \textbf{Output} & \textbf{Operation}\\
    \midrule
    Input Layer & (m, 32, 32, 3)  & (m, 32, 32, 64) & \textsc{Conv(3, 64, 3, 1)}, \textsc{LRelu}\\
    \midrule
    Hidden Layer & (m, 32, 32, 64)  & (m, 16, 16, 64) & \textsc{Conv(64, 64, 4, 2)}, \textsc{LRelu}\\
    Hidden Layer & (m, 16, 16, 64)  & (m, 16, 16, 128) & \textsc{Conv(64, 128, 3, 1)}, \textsc{LRelu}\\
    Hidden Layer & (m, 16, 16, 128)  & (m, 8, 8, 128) & \textsc{Conv(128, 128, 4, 2)}, \textsc{LRelu}\\
    Hidden Layer & (m, 8, 8, 128)  & (m, 8, 8, 256) & \textsc{Conv(128, 256, 3, 1)}, \textsc{LRelu}\\
    Hidden Layer & (m, 8, 8, 256)  & (m, 4, 4, 256) & \textsc{Conv(256, 256, 4, 2)}, \textsc{LRelu}\\
    Hidden Layer & (m, 4, 4, 256)  & (m, 4, 4, 512) & \textsc{Conv(256, 512, 3, 1)}, \textsc{LRelu}\\
    Hidden Layer & (m, 4, 4, 512)  & (m, 512) & \textsc{GSP}\\
    \midrule
    Output Layer & (m, 512)  & (m, 1) & \textsc{FC}(512, 1)\\
    \bottomrule
  \end{tabular}
  \caption{Discriminator of SNDCGAN~\citep{miyato2018spectral} used for CIFAR10 image synthesis.}
  \label{SNDCGAN_D}
\end{table*}

\begin{table*}[p]
  \centering
  \begin{tabular}{llrc}
    \toprule
    \textbf{Layer} & \textbf{Input} & \textbf{Output} & \textbf{Operation}\\
    \midrule
    Input Layer & (m, 128)  & (m, 16384) & \textsc{FC(128, 16384)}\\
    \midrule
    Reshape Layer & (m, 16384)  & (m, 4, 4, 1024) & \textsc{Reshape}\\
    Hidden Layer & (m, 4, 4, 1024) & (m, 8, 8, 512) & \textsc{Gen\_Block}(1024, 512, True) \\
    Hidden Layer & (m, 8, 8, 512) & (m, 16, 16, 256) & \textsc{Gen\_Block}(512, 256, True) \\
    Hidden Layer & (m, 16, 16, 256) & (m, 32, 32, 128) & \textsc{Gen\_Block}(256, 128, True) \\
    Hidden Layer & (m, 32, 32, 128) & (m, 64, 64, 64) & \textsc{Gen\_Block}(128, 64, True) \\
    
    Hidden Layer & (m, 64, 64, 64) & (m, 64, 64, 3) & \textsc{BN}, \textsc{ReLU}, \textsc{Conv(64, 3, 3, 1)} \\
    \midrule
    Output Layer & (m, 64, 64, 3)  & (m, 64, 64, 3) & \textsc{Tanh} \\
    \bottomrule
    
  \end{tabular}
  \caption{Generator of SNResGAN used for LSUN and iNaturalist-2019 image synthesis.}
  \label{SNResGAN_G}
\end{table*}
\begin{table*}[p]
  \centering
  \begin{tabular}{llrc}
    \toprule
    \textbf{Layer} & \textbf{Input} & \textbf{Output} & \textbf{Operation}\\
    \midrule
    Input Layer & (m, 64, 64, 3)  & (m, 32, 32, 64) & \textsc{Dis\_Block}(3, 64, True)\\
    \midrule
    Hidden Layer & (m, 32, 32, 64)  & (m, 16, 16, 128) & \textsc{Dis\_Block}(64, 128, True)\\
    Hidden Layer & (m, 16, 16, 128)  & (m, 8, 8, 256) & \textsc{Dis\_Block}(128, 256, True)\\
    Hidden Layer & (m, 8, 8, 256)  & (m, 4, 4, 512) &      \textsc{Dis\_Block}(256, 512, True)\\
    Hidden Layer & (m, 4, 4, 512)  & (m, 4, 4, 1024) & \textsc{Dis\_Block}(512, 1024, False), \textsc{ReLU}\\
    Hidden Layer & (m, 4, 4, 1024)  & (m, 1024) & \textsc{GSP}\\
    \midrule
    Output Layer & (m, 1024)  & (m, 1) & \textsc{FC(1024, 1)}\\
    \bottomrule
    
  \end{tabular}
    \caption{Discriminator of SNResGAN \citep{miyato2018spectral, gulrajani2017improved} used for LSUN and iNaturalist-2019 image synthesis.}
    \label{SNResGAN_D}
\end{table*}

\subsection{Hyperparameter Configuration (Image Generation Experiments)}

\subsubsection{Lambda  the Regularizer coeffecient}
The $\lambda$ hyperparameter is the only hyperparameter that we change
across different imbalance scenarios. As the overall objective is 
composed of the two terms:
\begin{equation}
    L_{g} = - E_{(x, z) \sim (P_r, P_z)}[\log(\sigma(D(G(z)) - D(x))] + \lambda L_{reg}
\end{equation}
As the number of terms in the regularizer objective can increase
with number of classes $K$. For making the regularizer term invariant of $K$
and also keeping the scale of regularizer term similar to GAN loss, we normalize it by $K$. Then the loss is multiplied by $\lambda$. Hence
the effective factor that gets multiplied with regularizer term is $\frac{\lambda}{K}$.

The presence of pre-trained classifier which provides labels for generated images makes it easy to determine the value of $\lambda$. Although the pre-trained classifier is trained on long-tailed
data its label distribution is sufficient to provide a signal for balance in generated distribution. We use the KL Divergence of labels with respect to uniform
distribution for $10$k samples in validation stage to check for balance in distribution and choose $\lambda$ accordingly.
We use the FID implementation available here \footnote{https://github.com/mseitzer/pytorch-fid}.

\subsubsection{Other Hyperparameters}
\begin{figure}[ht]
    \centering
    \includegraphics[width=\linewidth]{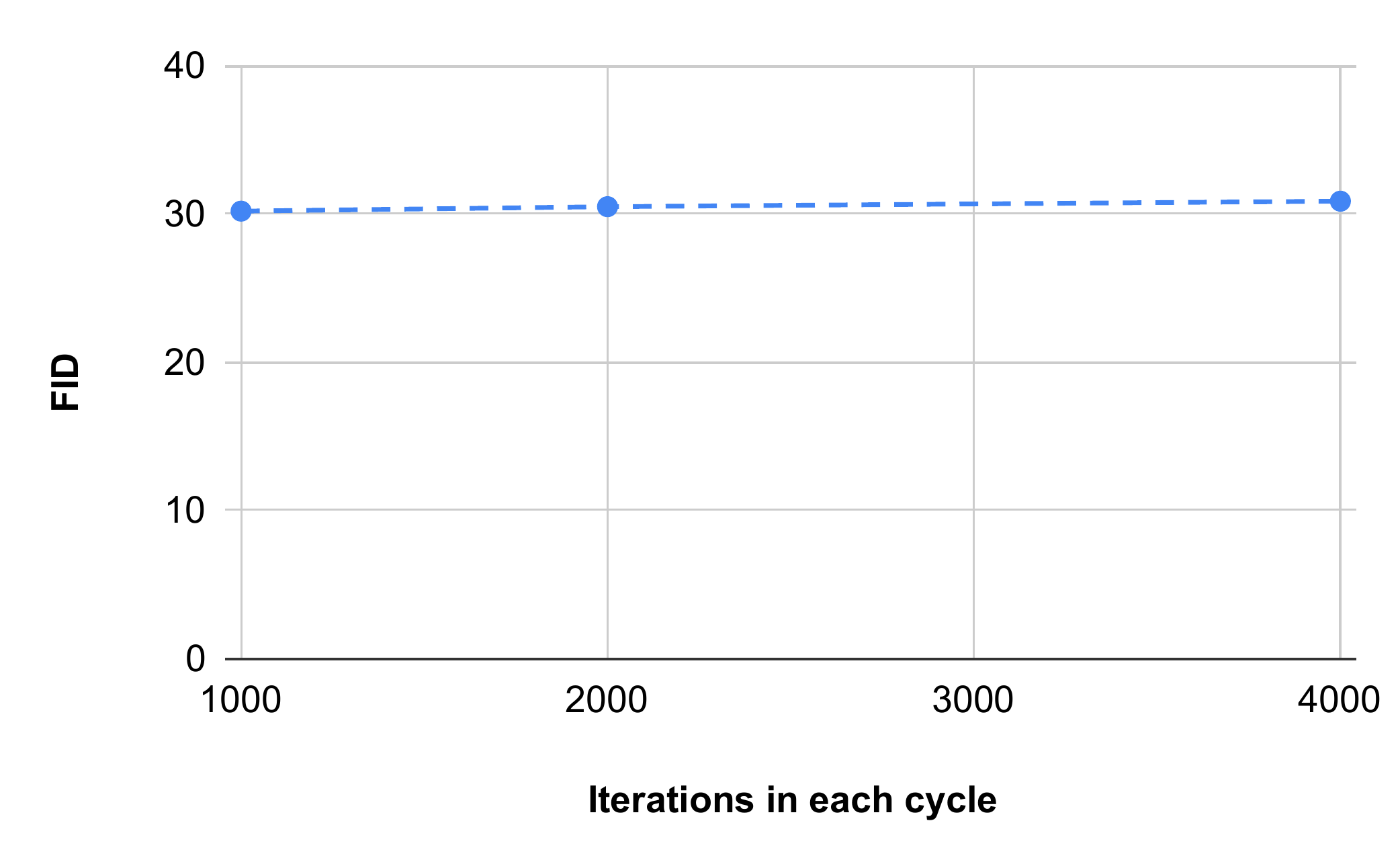}
    \caption{Effect on FID with change in number of steps in each cycle. After each cycle effective class statistics are updated (For CIFAR-10 imbalance ratio $\rho = 10$).}
    \label{fig:fid_stats}
\end{figure}

\begin{figure*}
\begin{subfigure}{.5\textwidth}
  \centering
  \includegraphics[width=.8\linewidth]{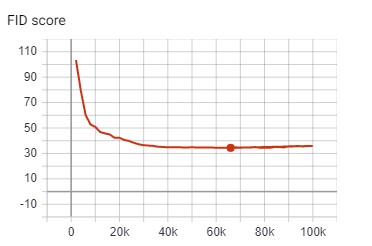}
   \caption{GAN}
  \label{fig:fid_gan}
\end{subfigure}%
\begin{subfigure}{.5\textwidth}
  \centering
  \includegraphics[width=.8\linewidth]{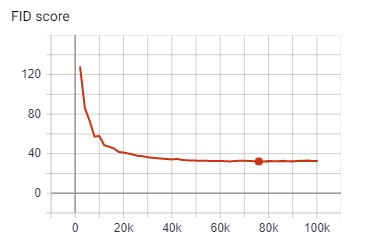}
  \caption{GAN + Proposed Regularizer}
  \label{fig:fid_our_gan}
\end{subfigure}
\caption{Plots of FID (y axis) vs Number of iteration steps. We observe a similar curve in both the cases.}
\label{fig:fid_curve}
\end{figure*}

We update the effective class distribution periodically after $2$k updates (i.e. each cycle defined in section \ref{sec:method} consists of 2k iteration). We find the algorithm performance to be stable for a large 
range of update frequency depicted in Figure \ref{fig:fid_stats}.
We also apply Exponential Moving Average on generator weights after $20$k steps for better generalization. The hyperparameters are present in detail in Table~\ref{tab:hyperparams}.
\\
\textbf{Validation Step:} We obtain the FID on 10k generated samples after each $2$k iterations and choose the checkpoint with best FID for final sampling and FID calculation present in
Table \ref{tab:gan_result}. 
\\
\textbf{Convergence of Network}: We find that our GAN + Regularizer setup also achieves similar convergence in FID value to the GAN without the regularizer. We show the FID curves for the CIFAR-10 (Imbalance Ratio = 10) experiments in Figure \ref{fig:fid_curve}. \\
\textbf{Ablation on $\beta$:} We find that for the CIFAR-10 dataset ($\rho = 10$) the choice of $\beta=1$ obtains similar FID ($30.48$) to $\beta=\alpha$ which obtains FID of $30.46$. The KL Divergence is also approximately the same for both cases i.e. $0.01$.

\begin{table}[h]
    \centering
    \begin{tabular}{lcc}
    \hline
         Parameter & Values(CIFAR-10) & Values(LSUN)\\ \hline
         Iterations & 100k & 100k\\ 
         $\beta$   & 1 & 1 \\
         Generator lr&  0.002&  0.002 \\ 
         Discriminator lr& 0.002 & 0.002\\ 
         Adam ($\beta_1$) & 0.5 & 0.0\\ 
         Adam ($\beta_2$) & 0.999 & 0.999\\ 
         Batch Size & 256 & 256\\ 
         EMA(Start After) & 20k & 20k\\ 
         EMA(Decay Rate) & 0.9999 & 0.9999\\ \hline
    \end{tabular}
    \caption{Hyperparameter Setting for Image Generation Experiments.}
    \label{tab:hyperparams}
\end{table}

\subsubsection{Hyperparameters for the Semi Supervised GAN Architecture}
\label{sub:semi-sup-hyper}
We use a ImageNet and ImageNet-21k pre-trained model with ResNet 50 architecture as the base model. The fine tuning of the model on CIFAR-10 and LSUN has been done by using the code of notebook present here
\footnote{https://github.com/google-research/big\_transfer/blob/master \\/colabs/big\_transfer\_pytorch.ipynb}. The accuracy of the classifiers fine-tuned on validation data, trained with $0.1\%$ of labelled data is 84.96 \% for CIFAR-10 and 82.40 \% for LSUN respectively.
The lambda (regularizer coefficient) values are present in the table below:

\begin{table}[h]
    \centering
    \begin{tabular}{lccc}
        \hline
        Imbalance Ratio ($\rho$) & 100  &  10 \\ \hline
        CIFAR-10 & 10 & 7.5\\ 
        LSUN & 10 & 7.5 \\ \hline
        
    \end{tabular}
    \caption{Values of $\lambda$ for different imbalance cases. For LSUN the $\lambda$ gets divide by 5 and for $\lambda$ it gets divided by 10 before multiplication to regularizer term.}
    \label{tab:my_label}
\end{table}

The training hyper parameters are same as the ones present in the Table \ref{tab:hyperparams}. Only in case of LSUN semi supervised experiments we use a batch size of 128 to fit into GPU memory for semi supervised experiments. 

\label{app: hyperparams}

\subsection{Experimental Details on CIFAR-100 and INaturalist-2019}
\label{app:largedatasets}
\subsubsection{CIFAR-100}

In this section we show results on CIFAR-100 dataset which has 100 classes having 500 training images for each class. We use SNResGAN architecture from \cite{miyato2018spectral} for generating $32 \times 32$ size images, which is similar to SNResGAN  architecture used for LSUN experiments. We use the same hyper-parameters used for LSUN experiments listed in Table \ref{tab:hyperparams}. We use a $\lambda$ value of 0.5 for CIFAR-100 experiments, the effective value that will get multiplied with $L_{reg}$ is $\frac{0.5}{100}$. The results in Table \ref{tab:cifarinat} show that our method on long-tailed CIFAR100 of using GAN + Regularizer achieves the best FID and also have class balance similar to cGAN (conditional GAN). The labels for the samples generated by GAN are obtained by a classifier trained on balanced CIFAR-100 dataset for KL Divergence calculation. The KL Divergence between the GAN label distribution and uniform distribution is present in Table \ref{tab:cifarinat}. The classifier for obtaining class labels for KL Divergence evaluation is trained on balanced CIFAR-100 with setup described in~\ref{sec:pretrained_classifier} which serves as annotator for all methods. The balanced classifier achieves an accuracy of 70.99\% and the pre-trained classifier trained on long-tailed data achieves an accuracy of 57.63\%. The pre-trained classifier is used in the process of GAN training. 

\subsubsection{iNaturalist-2019}
We use SNResGAN architecture described in Table \ref{SNResGAN_D} and Table \ref{SNDCGAN_G} for generating $64 \times 64$ images for the iNaturalist 2019 dataset. In case of iNaturalist all batch-norms are conditional batch norms (cBN) in Generator, in case of our method and the unconditional baseline (SNResGAN) we use random labels as conditioning labels. As in our method have access to class distribution $N_k^t$ we sample random labels with weight distribution proportional to $1/N_k^t$. With this change we see an FID decrease from 11.58 to 9.01. The $\lambda$ value of 0.5 and $\alpha = 0.005$ is used for the experiments. The effective value of $\lambda$ that will be multiplied with $L_{reg}$ is $\frac{0.5}{1010}$. The statistics $N_k$ is updated for each iteration in this case. We follow SAGAN \citep{pmlr-v97-zhang19d} authors recommendation and use spectral normalization in generator in addition to the discriminator for stability on large datasets.  Other hyper-parameters are present in Table \ref{tab:hyperparams_cifarinat}.

\begin{table}[h]
    \centering
    \begin{tabular}{lcc}
    \hline
         Parameter & Values(CIFAR-100) & Values(iNat19)\\ \hline
         Iterations & 100k & 200k\\ \
         $\beta$   & 1 & $\alpha$ \\ \
         Generator lr&  0.002&  0.002 \\ \
         Discriminator lr& 0.002 & 0.002\\ \
         Adam ($\beta_1$) & 0.5 & 0.0\\ \
         Adam ($\beta_2$) & 0.999 & 0.999\\ \
         Batch Size & 256 & 256\\ \
         EMA(Start After) & 20k & 20k\\ \
         EMA(Decay Rate) & 0.9999 & 0.9999\\ \hline
    \end{tabular}
    \caption{Hyperparameter Setting for Image Generation Experiments on CIFAR-100 and iNatuaralist-2019}
    \label{tab:hyperparams_cifarinat}
\end{table}
The pre-trained classifier for iNaturalist-2019 is ResNet-32 trained with usual cross entropy loss with learning schedule as described in Appendix~\ref{sec:pretrained_classifier}. The classifier is trained on resolution of $224 \times 224$ and achieves an accuracy of 46.90 \% on the validation set. As in iNaturalist 2019 case we don't have balanced training set available we use this classifier to get the KL Divergence from uniform distribution.

\subsection{Pre-trained Classifier Baseline}
\label{app:other_baseline}
\begin{figure*}[ht]
\centering
\begin{subfigure}{0.5\columnwidth}
   \centering
  \includegraphics[width=0.9\columnwidth]{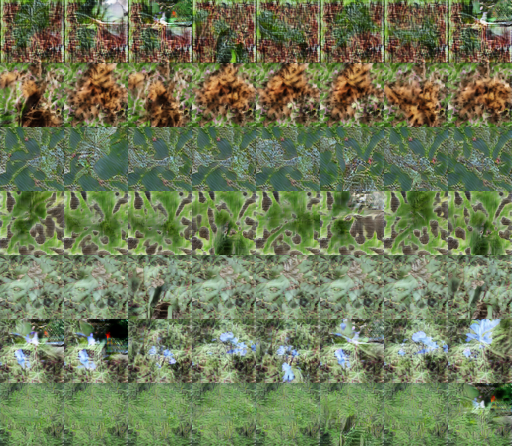}
  \caption{Classifier \\ Baseline}
\end{subfigure}%
\begin{subfigure}{0.5\columnwidth}
   \centering
  \includegraphics[width=0.9\columnwidth]{figures/inat_ours.png}
  \caption{Our \\ method}
\end{subfigure}

\caption{Qualitative comparison of images generated by ACGAN like baseline and our method.}
\label{fig:other_baseline_images}
\end{figure*}

In these experiments we introduce a pre-trained classifier in place of the classifier being jointly learned by ACGAN on iNaturalist-2019 dataset. This makes the comparison fair as both approaches use a pre-trained classifier and unlabelled images. The hyper-parameters used are same as present in Table \ref{tab:hyperparams_cifarinat}. We use a classification loss (i.e. cross entropy loss) in addition to the GAN loss similar to ACGAN. The $\lambda$ value for the classification loss is set to $0.5$. The illustration of the approach is present in the Figure \ref{fig:approach_overview}. We find that this approach leads to mode collapse and is not able to produce diverse samples within each class. The comparison of the generated images is present in the Figure \ref{fig:other_baseline_images}. \\

\begin{figure*}[h]
\centering
\begin{subfigure}{.4\textwidth}
  \centering
  \includegraphics[width=.8\linewidth]{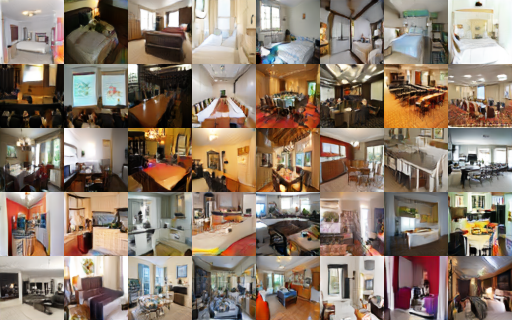}  
  \caption{ACGAN (Conditional)}
  \label{fig:sub-first}
\end{subfigure}
\begin{subfigure}{.4\textwidth}
  \centering
  \includegraphics[width=.8\linewidth]{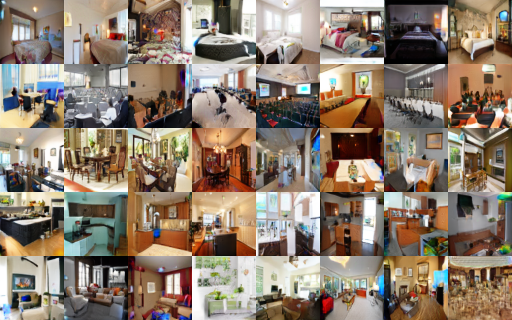}  
  \caption{cGAN (Conditional)}
  \label{fig:sub-second}
\end{subfigure}

\begin{subfigure}{.4\textwidth}
  \centering
  \includegraphics[width=.8\linewidth]{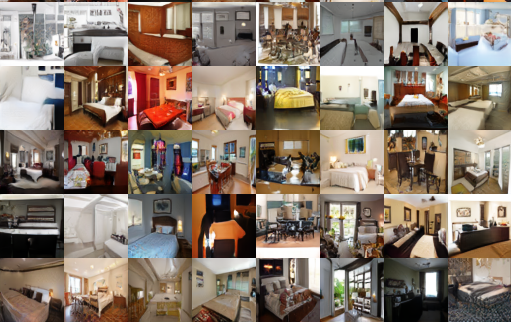}  
  \caption{SNResGAN (Unconditional)}
  \label{fig:sub-third}
\end{subfigure}
\begin{subfigure}{.4\textwidth}
  \centering
  \includegraphics[width=.8\linewidth]{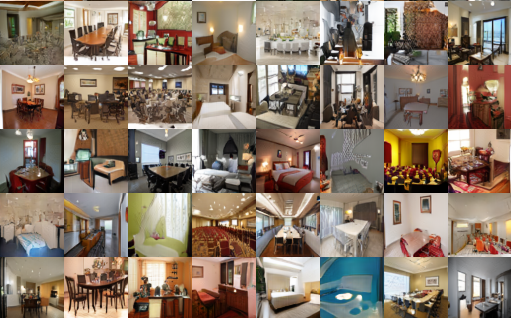}  
  \caption{Ours (Unconditional)}
  \label{fig:sub-fourth}
\end{subfigure}
\caption{Images from different GANs with imbalance ratio ($\rho = 10$)}
\label{fig:fig}
\end{figure*}

\begin{figure*}[h]
\centering
\begin{subfigure}{.4\textwidth}
  \centering
  \includegraphics[width=.8\linewidth]{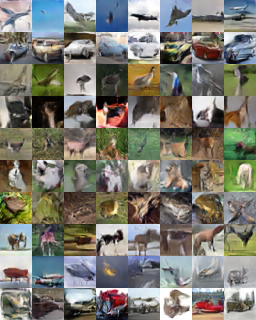}  
  \caption{ACGAN (Conditional)}
  \label{fig:sub-first}
\end{subfigure}%
\begin{subfigure}{.4\textwidth}
  \centering
  \includegraphics[width=.8\linewidth]{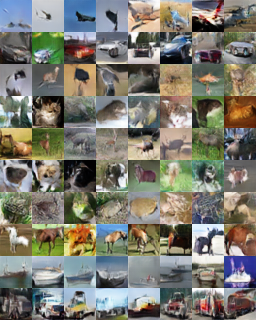}  
  \caption{cGAN (Conditional)}
  \label{fig:sub-second}
\end{subfigure}

\begin{subfigure}{.4\textwidth}
  \centering
  \includegraphics[width=.8\linewidth]{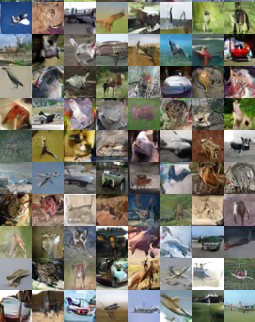}  
  \caption{SNDCGAN (Unconditional)}
  \label{fig:sub-third}
\end{subfigure}
\begin{subfigure}{.4\textwidth}
  \centering
  \includegraphics[width=.8\linewidth]{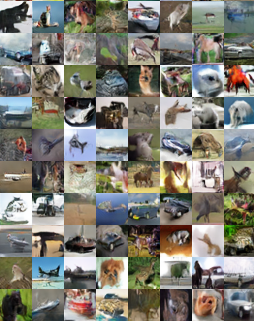}  
  \caption{Ours (Unconditional)}
  \label{fig:sub-fourth}
\end{subfigure}
\caption{Images generated by different GANs for CIFAR-10 with imbalance ratio ($\rho = 10$).}
\label{fig:fig}
\end{figure*}

\begin{figure*}[h]
    \centering
    \includegraphics{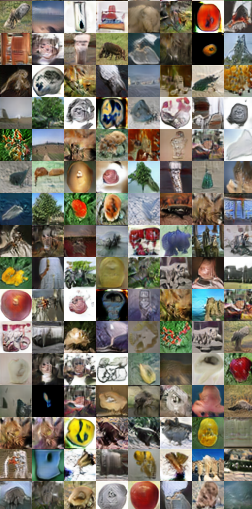}
    \caption{Images generated for CIFAR-100 dataset with our method (GAN + Regularizer).}
    \label{fig:cifar100}
\end{figure*}

\end{document}